%% file: main.tex
\documentclass[final,3p,times,twocolumn]{elsarticle}

\usepackage{xcolor,colortbl}
\usepackage{amssymb}
\usepackage{amsmath}
\usepackage{amsfonts}
\usepackage{soul, color}
\usepackage{wrapfig}

\newcommand{\ve}[1]{\mathbf{#1}} 
\newcommand{\tve}[1]{\tilde{\mathbf{#1}}} 
\newcommand{\bve}[1]{\mathbf{\bar{#1}}} 

\journal{~}

\begin{document}

\begin{frontmatter}



\title{Learning Human-arm Reaching Motion Using IMU\\ in Human-Robot Collaboration}

\author[label1]{Nadav D. Kahanowich and Avishai Sintov\corref{cor1}}
\affiliation[label1]{organization={School of Mechanical Engineering, Tel-Aviv University},
            addressline={Haim Levanon St.},
            city={Tel-Aviv},
            postcode={6997801},
            country={Tel-Aviv}}
\ead{kahanowich@mail.tau.ac.il, sintov1@tauex.tau.ac.il}
\cortext[cor1]{Corresponding Author.}
%





\begin{abstract}
Many tasks performed by two humans require mutual interaction between arms such as handing-over tools and objects. In order for a robotic arm to interact with a human in the same way, it must reason about the location of the human arm in real-time. Furthermore and to acquire interaction in a timely manner, the robot must be able predict the final target of the human in order to plan and initiate motion beforehand. In this paper, we explore the use of a low-cost wearable device equipped with two inertial measurement units (IMU) for learning reaching motion for real-time applications of Human-Robot Collaboration (HRC). A wearable device can replace or be complementary to visual perception in cases of bad lighting or occlusions in a cluttered environment. We first train a neural-network model to estimate the current location of the arm. Then, we propose a novel model based on a recurrent neural-network to predict the future target of the human arm during motion in real-time. Early prediction of the target grants the robot with sufficient time to plan and initiate motion during the motion of the human. The accuracies of the models are analyzed concerning the features included in the motion representation. Through experiments and real demonstrations with a robotic arm, we show that sufficient accuracy 
is achieved for feasible HRC without any visual perception. Once trained, the system can be deployed in various spaces with no additional effort. The models exhibit high accuracy for various initial poses of the human arm. Moreover, the trained models are shown to provide high success rates with additional human participants not included in the model training.
\end{abstract}




\end{frontmatter}

\section{Introduction}
\label{sec:introduction}
\input{introduction}

\section{Related Work}
\label{sec:relatedwork}
\input{related_work}

\section{Method}
\label{sec:method}
\input{method}

\section{Experiments}
\label{sec:experiments}
\input{experiments}

\section{Conclusion}
\label{sec:conclusion}

\input{conclusions}

\section*{Declaration of competing interest}

The authors declare that they have no known competing financial interests or personal relationships that could have appeared to influence the work reported in this paper.

\section*{Funding}

This work was supported by the Israel Science Foundation (grant No.
1565/20).

 \bibliographystyle{elsarticle-num} 
 \bibliography{ref}








\end{document}

%% file: introduction.tex
When two humans perform a shared task, each has an ability to predict intentions of his peer without verbal communication. Once one human sees the motion of his human fellow, usually his arms and manipulated objects, the intended upcoming task can be predicted for further interaction \cite{Rizzolatti2001}. For instance, when a human is handing-over an object, his fellow can infer about the reaching target and initiate supporting motion to obtain it. In \textit{Human-Robot Collaboration} (HRC), a robot should act the same to support a human in completing shared tasks \cite{Ajoudani2017}. Having a robot infer about human upper-limbs motion has applications in hand-over activities \cite{Yang2021}, collaboration in shared workspaces \cite{Luo2018}, collision avoidance by the robot \cite{Lasota2017} and virtual reality \cite{Gamage2021}.
\begin{figure}
    \centering
    \includegraphics[width=\linewidth]{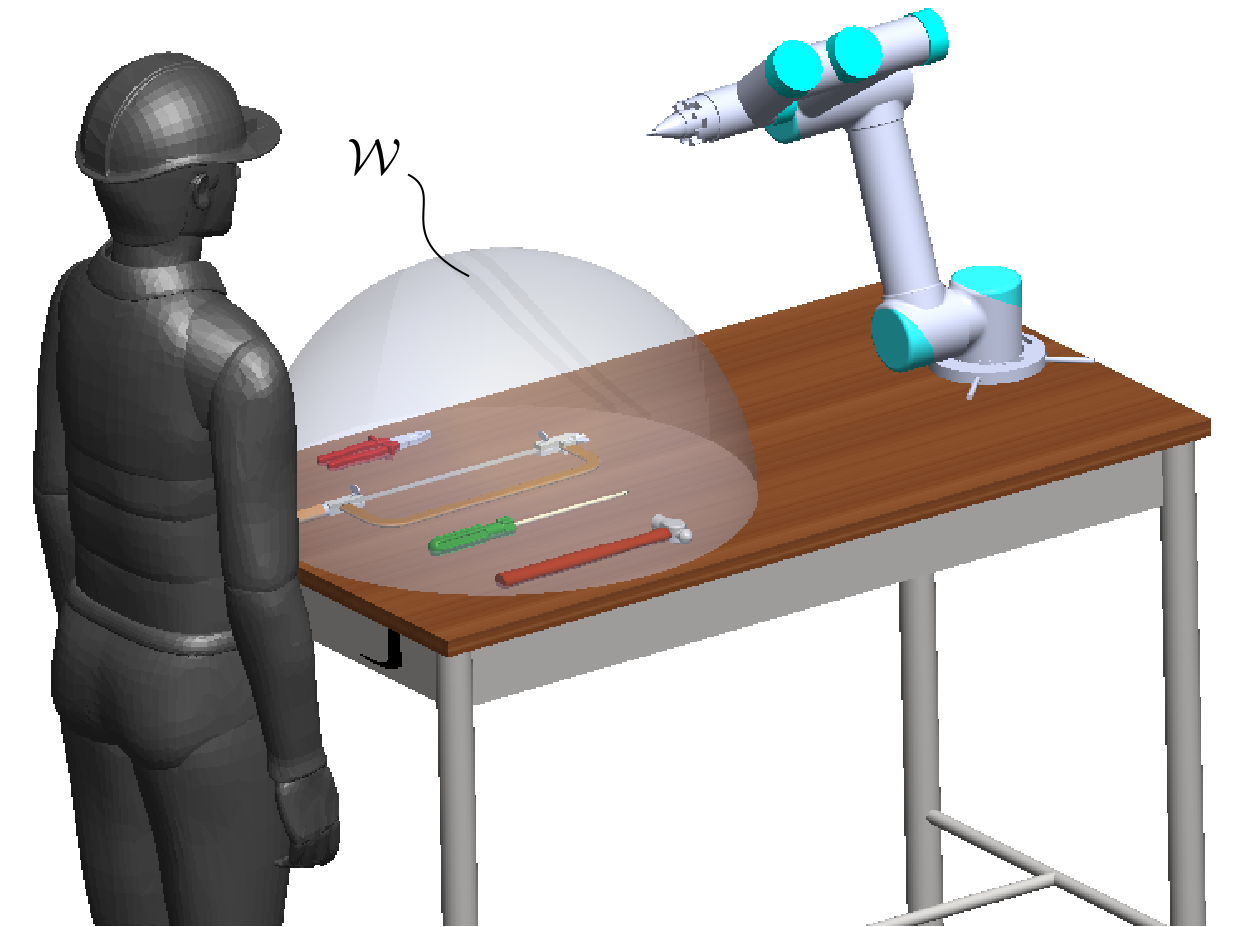}
    \caption{Human user and robot share space $\mathcal{W}$ in which they are to interact or collaborate in a shared task.}
    \label{fig:W}
\end{figure}

Human motion inference and prediction have been given much attention in the past few decades. Seminal work by Flash and Hogen \cite{Flash1985} has suggested that human arms move from point to point in a smooth trajectory while minimizing the mean-square jerk. Others proposed models that include minimum torque \cite{Uno1989} and position variances \cite{Harris1998}. Nevertheless, these models were demonstrated in a limited and specific setting while human motion is difficult to predict due to the randomness and complexity of human behavior \cite{Peng2014}. Therefore, a significant amount of research has been put on the use of Kalman filter variants \cite{Kohler1997,Yun2006} and data-driven models \cite{Wu2014, Cheng2019}. In an example for the latter, the work of Landi et al. \cite{Landi2019} combined the minimum jerk model with an Artificial Neural-Network (ANN) to predict arm motion based on camera perception. However, all above methods rely on visual perception in order to acquire human arm pose. 
Relying on continuous visual feedback limits the performance of various tasks in which visual uncertainty (e.g., poor lighting or shadows) or occlusion may occur. Moreover, dealing with visual sensing requires a large amount of data and strong computing capabilities \cite{Toshev2014}.

To bypass the challenges of vision, motion prediction using wearable sensors has also been exhibited. For instance, Electro-Myography (EMG) \cite{Bi2019} and brain-computer interface \cite{Nakanishi2013} have been tested for intention prediction of motor behaviour. Wearable Inertial Measurement Units (IMU) have also been proposed and yielded significant results \cite{Corrales2008}. The work in \cite{Tian2015} fused IMU measurements with depth camera (Kinect) perception based on the Unscented Kalman Filter (UKF). Yun and Bachmann \cite{Yun2006}, on the other hand, filtered two IMU sensors on the upper-arm and forearm to approximate the orientation of the arm. Similarly, Atrsaei et al. \cite{Atrsaei2018} used two IMU sensors on the upper-arm and forearm along with UKF to approximate the pose of the arm. 
However, the methods were focused on approximating the current pose of the arm and did not consider future arm trajectories. In addition, the ability of the above methods to generalize to various participants has not been demonstrated. 

In this paper, we aim to rely solely on a low-cost wearable device directly strapped on the human arm to provide an affordable solution. Achieving vision-free pose estimation and prediction would enable further fusion with vision for better estimation in unstructured or partly-occluded environments. Therefore, we explore the sole use of IMU for learning reaching motion of a human arm. With two IMU sensor located on the upper-arm and forearm, we observe the required data and learning model to predict the target of the reaching arm at early stages of motion in real-time. A learning approach is proposed based on the Long Short Term Memory (LSTM) model. We investigate the use of raw data from the IMU's along with temporal pose predictions of the arm acquired from a secondary ANN model. Therefore, our proposed method does not require integrating the measured accelerometers in order to predict the position. Integration is often subject to drift and sensitive to typical noisy signals \cite{Tian2015}. Also and contrary to the common approach, our method does not require the use of filters nor early calibration of the wearable device. Hence, it can be used immediately upon placement. We also observe the robustness of the method to taking-off and re-positioning the device on the arm, and for several users. 

The main contribution is an approach that enables real-time target prediction in reaching motions of a human arm using affordable, lightweight and easy to use hardware, and without the need for visual perception. With early real-time information, the robot could promptly plan and initiate a response motion in order to interact in a timely manner. The approach is also shown to be able to cope with variance in the pose of the human user standing in front of the robot. While we focus on collaboration with robotic arms, the approach can also be applicable to interactions with prosthetic hands, drones, tele-operation and virtual reality. 

The remainder of this paper is organized as follows. Section \ref{sec:relatedwork} discusses relevant literature. Section \ref{sec:method} establishes the addressed problem, the proposed system and learning method. Furthermore, Section \ref{sec:experiments} presents the data collection, experiments, analysis and demonstration for reaching prediction. Finally, Section \ref{sec:conclusion} concludes the paper.

%% file: related_work.tex
As described in Section \ref{sec:introduction}, a major part of prior work focused on the use of visual perception in order to investigate human arm motion. As such, Oguz et al. \cite{Oguz2018} solved an inverse optimal control problem in order to derive the true cost function that governs a set of motions. The work in \cite{Tamura2006} predicted to which object the human hand is reaching. The target object was predicted by observing arm motion with a camera in a discretized workspace. It is also worth mentioning extensive attempts for pose and motion prediction of the entire human body through visual perception \cite{Fragkiadaki2015,Martinez2017,Salzmann2020}. Similarly, depth camera such as Kinect is an alternative solution using spatial point cloud \cite{Wu2018}. 

While the above focused on extracting mathematical models of motion, a different approach involves data-driven models. The work of Cheng et al. \cite{Cheng2019} proposed the use of semi-adaptable ANN to learn a human arm transition model and adapt it to time-varying human behaviors. Another approach uses Hidden Markov Models to approximate human pose or occupied workspace based on visual observations \cite{Ding2011,Wu2014}.  The use of Recurrent Neural-Networks (RNN) is also a common approach where sequential temporal data is used to predict motion trajectories \cite{Liu2019,Zhang2020}. Liu and Liu \cite{Liu2021} combined a Modified Kalman Filter to adapt an RNN model to changes in environmental conditions. The work in \cite{DArpino2015} used a database of recorded human motions in order to predict in real-time intended targets in reaching motions. However and as previously indicated, relying solely on continuous visual feedback limits the performance in unstructured environments where occlusions may occur. Therefore, in this work we focus on observing human arm motion through wearable sensors while later fusion with visual perception may provide a complete solution.

Object hand-over from a robot to a human, and vice versa, is a potential application which requires reasoning about human arm pose and future motion. Visual perception is the leading method to locate the human arm and approximate its pose \cite{Rosenberger2021}. In \cite{Yang2020}, human-to-robot handovers were conducted by visually classifying a human grasp of the reached-out object and planning an approach trajectory for the robot. Nemlekar et al. \cite{Nemlekar2019} approximated the object transfer point based on observed human behavior and motion.

%% file: method.tex
\subsection{Problem formulation}

A human user stands in front of a designated interaction-space $\mathcal{W}\in\mathbb{R}^3$. Interaction-space $\mathcal{W}$ is shared with a robotic arm such that both user and robot can reach all points within it (Figure \ref{fig:W}). Let the position of the human wrist at time $t$ be $\ve{p}(t)\in\mathbb{R}^3$ relative to some coordinate frame on $\mathcal{W}$. At time $t=0$, the wrist is at some arbitrary pose. The initial pose is randomly selected close to the body of the user (e.g., side of the hip or palm on the chest).
The user then moves his arm to reach some point $\ve{p}(t_f)\in\mathcal{W}$ where $t_f<T$ for a pre-defined upper-bound $T$.

Let $\ve{x}(t)\in\mathbb{R}^n$ be the state vector of some $n$ measurable features on the human arm and denote $\tve{p}(t)\in\mathbb{R}^3$ to be the estimated position of the wrist at time $t$. In addition, we denote $\tve{p}_f(t)\in\mathcal{W}$ to be the estimated destination of the wrist approximated at time $t<T$. We set two goals. First, we aim to localize the human wrist in real-time during motion. Hence, we search for a map $\Gamma:\mathbb{R}^n\to\mathbb{R}^3$ that approximates the wrist position $\tve{p}(t)=\Gamma(\ve{x}(t))$ at time $t$. Second, we aim to learn a map $\Phi:\mathcal{C}\to\mathcal{W}$ that provides a prediction at time $t$ of the expected wrist target at some time $t_f$. $\mathcal{C}$ is some unknown space based on measurable features in $\ve{x}(t)$ that can provide accurate prediction. In this work, we explore the formulation of $\mathcal{C}$ to acquire a sufficient representation of $\tve{p}_f(t)=\Phi(\ve{d}(t))$ for $\ve{d}(t)\in\mathcal{C}$.

\subsection{System}
\label{sec:system}

\begin{figure}
    \centering
    \includegraphics[width=0.7\linewidth]{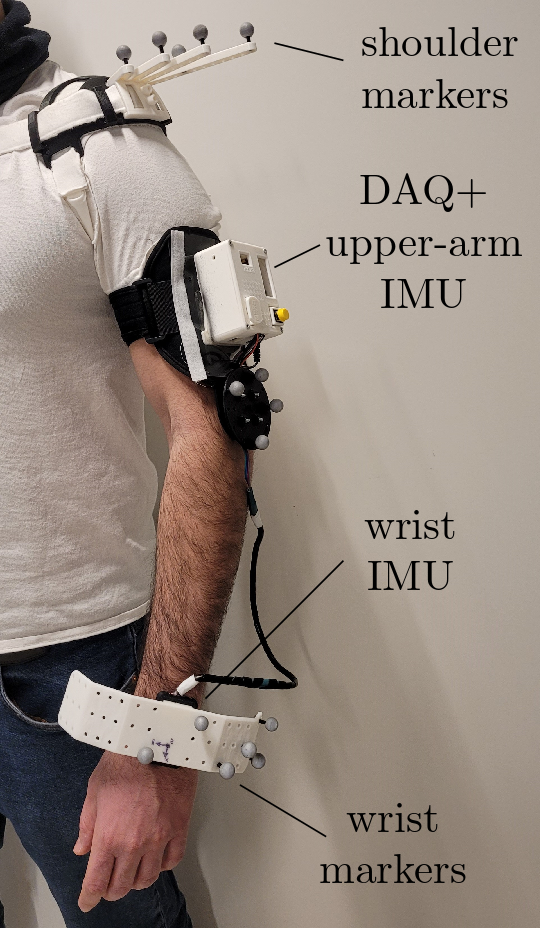}
    \caption{System of two IMU positioned on the wrist and upper-arm along with three sets of markers on the shoulder, upper-arm (not used) and wrist, observed by a motion capture system.}
    \label{fig:system}
\end{figure}

We have designed and fabricated an experimental wearable device consisting of three main components: (a) wrist band with IMU and reflective markers, (b) upper-arm band with IMU and (c) a data acquisition (DAQ) system based on an Arduino Uno board. We have also included shoulder and upper-arm reflective markers for analysis and verification while not used in the modeling. The system is seen in Figure \ref{fig:system}. Each IMU provides 3-axes of acceleration, angular change rate and orientation using accelerometer, gyroscope and a magnetometer, respectively. Hence, two IMU's provide a state vector $\ve{x}$ with maximum size of $n=18$. 

Measurements of a magnetometer are dependent on the relative orientation to earth's magnetic field. Without loss of generality, we consider an interaction-space at the same orientation relative to the standing position of the human. Hence, we assume that all recordings are taken within a limited orientation range of the human torso relative to the interaction space. Nevertheless, simply having another IMU on the human torso, magnetometer in particular, would enable relative measurements and additional interaction-spaces around the user.

The acquisition configuration provides real-time data stream of all the given sensors in a frequency of 60~Hz. The reflective markers  are tracked using a motion capture system and provide the positions of the shoulder, upper-arm and wrist bands. We note that the marker set is used only for data collection and validation while not required in the eventual system usage. Hence, the described system is composed of low-cost and light-weight (325~g) hardware which is appealing and suited for easy arm movements. 

\subsection{Data collection and formation}

Data is collected over a set of $K$ episodes. In each episode, the user starts from wrist position $\ve{p}(0)$ and moves the arm to final wrist position $\ve{p}(t_f)\in\mathcal{W}$. During motion of episode $j$, sets of states $\mathcal{X}_j=\{\ve{x}_j(0),\ldots,\ve{x}_j(t_f)\}$ and wrist positions $\mathcal{P}_j=\{\ve{p}_j(0),\ldots,\ve{p}_j(t_f)\}$ are recorded. For training model $\Gamma$, data from all episodes is pre-processed to have a labeled training dataset $\mathcal{M} = \{ (\ve{x}_i,\ve{p}_i)\}_{i=1}^{N}$. The formation of the training data for target prediction will be discussed later.

\subsection{Wrist position model}
\label{sec:wrist_position}

Let $\theta_{elv}$ and $\theta_{yaw}$ be elevation and yaw angles of the upper-arm, respectively. Similarly, let $\phi_{elv}$ and $\phi_{yaw}$ be the forearm elevation and yaw angles, respectively. The position of the wrist $\ve{p}=(p_x,p_y,p_z)^T$ is acquired by forward kinematics as given by Soechting and Flanders \cite{Soechting1989}:
\begin{align}
    p_x &= l_u\sin\theta_{elv}\sin\theta_{yaw} + l_f\sin\phi_{elv}\sin\phi_{yaw} \label{eq:p_x} \\ 
    p_y &= l_u\sin\theta_{elv}\cos\theta_{yaw} + l_f\sin\phi_{elv}\cos\phi_{yaw} \\
    p_z &= -l_u\cos\theta_{elv} + l_f\cos\phi_{elv}. \label{eq:p_z}
\end{align}
where $l_u$ and $l_f$ are the lengths of the upper-arm and forearm, respectively. Assuming $l_u$ and $l_f$ are known, expressions \eqref{eq:p_x}-\eqref{eq:p_z} show that the position of the wrist can be acquired by measuring the orientations of the upper-arm and forearm. While these orientations can be measured by the accelerometers of the two IMU's when in a static pose, the two magnetometers can also do so during arm motion. Nevertheless, acceleration can add viable information 
for reaching $\mathcal{W}$.

Given training dataset $\mathcal{M}$, we train a feed-forward ANN to obtain map $\tve{p}_j=\Gamma_\theta(\ve{x}_j)$. Vector $\theta$ consists of the trained parameters of the model. In such case, user lengths $l_u$ and $l_f$ are embedded in $\mathcal{M}$ and explicit measurements are not required. Furthermore, the features to be included in state $\ve{x}_j$ and its size $n$ such that $\Gamma_\theta$ achieves highest accuracy would be analyzed in the experimental section.


\subsection{Long-Short Term Memory (LSTM)}

LSTM is a class of Recurrent Neural-Networks (RNN) aimed to learn sequential data \cite{Yu2019}. RNN utilizes previous outputs as inputs while including hidden states. For each time step $t$, the hidden state vector $\ve{h}(t)$ and the output $\ve{y}(t)$ are expressed as
\begin{equation}
    \ve{h}(t) = g_1(W \ve{h}(t-1) + U \ve{x}(t) + \ve{b}_h)
\end{equation}
and
\begin{equation}
    \ve{y}(t) = g_2(V \ve{h}(t) + \ve{b}_y)
\end{equation}
where $W, U, V$ are weight matrices and $\ve{b}_h,\ve{b}_{y}$ are bias vectors. $g_1$ and $g_2$ are activation functions. The standard RNN is usually not capable of handling long intervals where back-propagating errors tend to vanish or explode \cite{Bengio1994}. LSTM, on the other hand, is capable of learning long-term dependencies by utilizing memory about previous inputs for an extended time duration \cite{hochreiter1997long}. 
Along with an hidden state vector, LSTM maintains a cell state vector $\ve{c}(t)$. At each time step, the process may choose to read from $\ve{c}(t)$, write to it or reset the cell using an explicit gating mechanism. Each cell unit has three gates of the same shape. The input gate controls whether the memory cell is updated or, in other words, which information will be stored. An LSTM cell can be formulated with the following expressions:
\begin{align}
    i(t) &= \sigma(W_i[\ve{h}(t-1), \ve{x}(t)] + \ve{b}_i)\\
    f(t) &= \sigma(W_{f}[\ve{h}(t-1), \ve{x}(t)] + \ve{b}_{f})\\
    o(t) &= \sigma(W_{o}[\ve{h}(t-1), \ve{x}(t)] + \ve{b}_{o})
\end{align}
where $W_i$, $W_f$ and $W_o$ are weight matrices. $\ve{b}_i$, $\ve{b}_f$ and $\ve{b}_o$ are bias vectors. Forget gate $f(t)$ controls whether the memory cell is reset and removes irrelevant information from the cell state. Similarly, output gate $o(t)$ controls whether the information of the current cell state is made visible and adds useful information to the cell state. Both gates have a Sigmoid activation function $\sigma$. To modify the cell state, another vector $\tilde{\ve{c}}(t)$ is defined as
\begin{equation}
    \tilde{\ve{c}}(t) = \tanh(W_{c}[\ve{h}(t-1), \ve{x}(t)] + \ve{b}_{c})
\end{equation}
where $W_c$ and $\ve{b}$ are weight matrix and bias vector, respectively. The hyperbolic activation function $\tanh{}$ distributes gradients and, therefore, prevents vanishing or exploding gradients, and allows a cell state information to flow longer. Vector $\tilde{\ve{c}}(t)$ is a new candidate that can be applied to the cell state in case the forget state chooses to reset. Hence, the new cell state $\ve{c}(t)$ is updated with
\begin{equation}
    \ve{c}(t) = f(t) \ve{c}(t-1) + i(t)\tilde{\ve{c}}(t).
\end{equation}
Furthermore, the hidden state is updated according to 
\begin{equation}
    \ve{h}(t) = \tanh(\ve{c}(t)) \cdot o(t).
\end{equation}
The LSTM is trained using recorded data sequences with back-propagation.
\begin{figure*}
    \centering
    \begin{tabular}{cc}
       \includegraphics[width=0.49\textwidth]{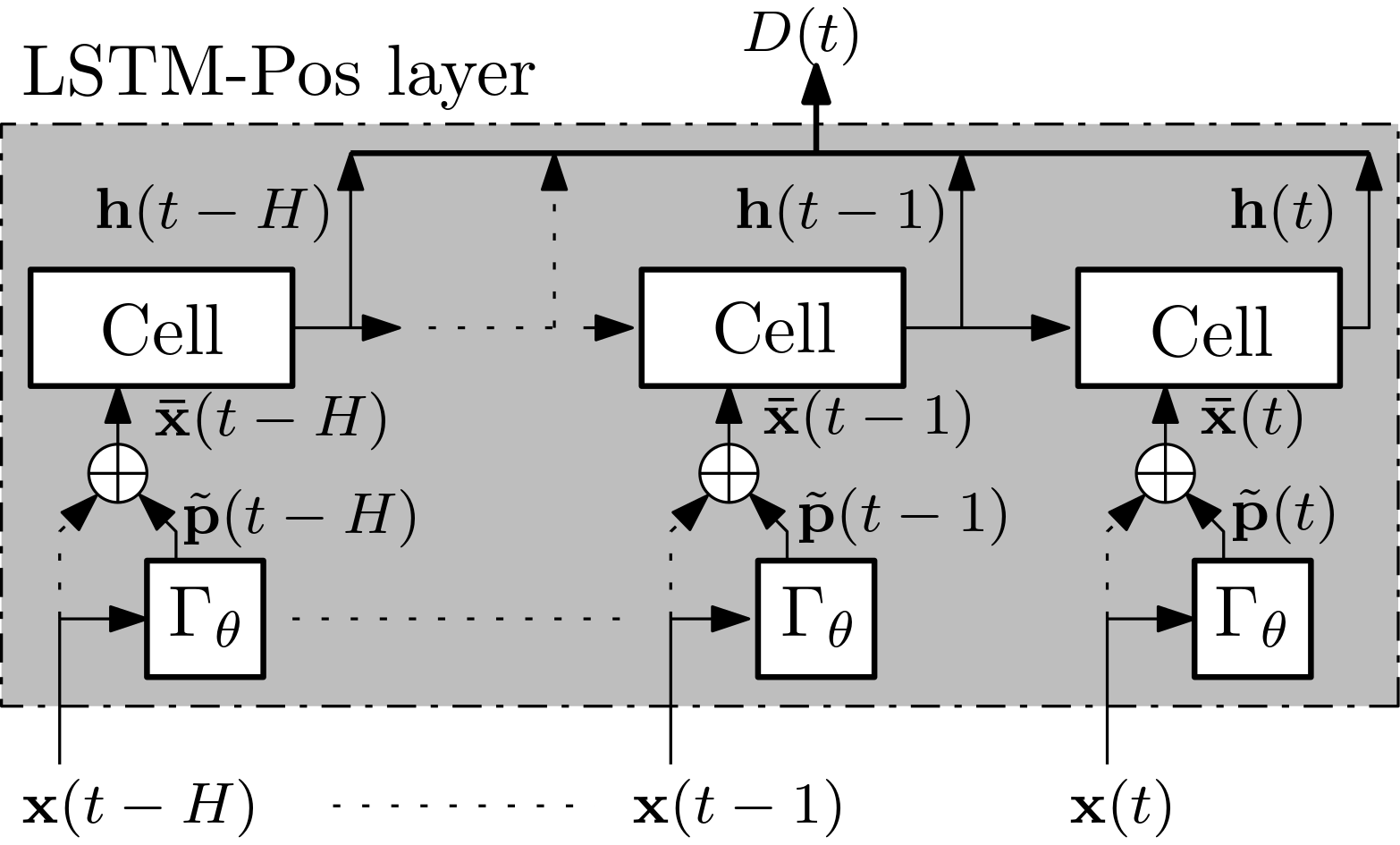}  &  \includegraphics[width=0.49\textwidth]{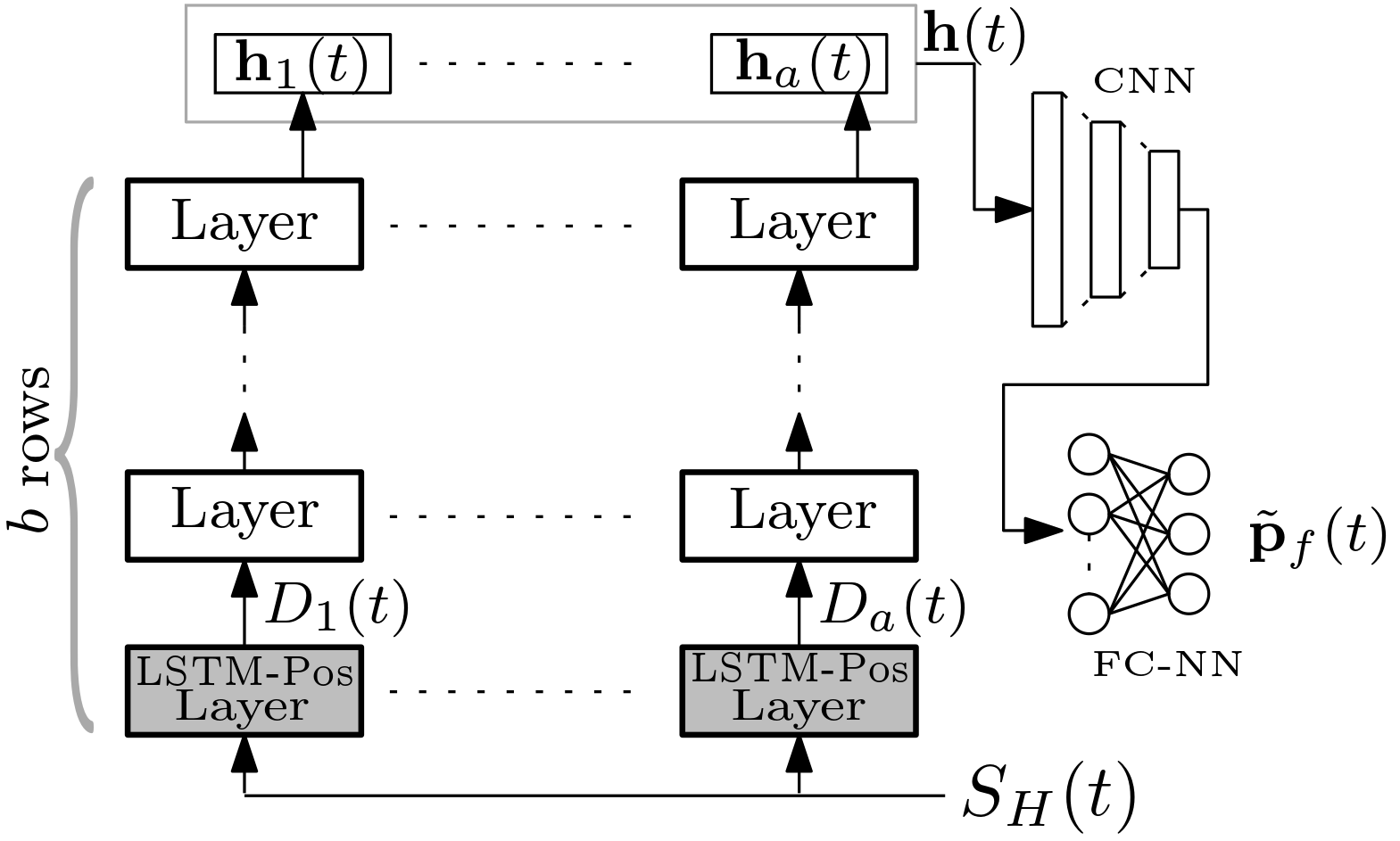}\\
       (a) & (b)
    \end{tabular}
    \caption{(a) A scheme of an LSTM-Pos layer receiving a state sequence $S_H(t)=\{\ve{x}(t-H),\ldots,\ve{x}(t)\}$ of length $H$ with output vector $\ve{h}_j(t)$. The requirement for including the raw data with concatenation (denoted with symbol $\oplus$) would be further analysed. (b) The proposed LSTM-Pos network with stacked layers followed by a feed-forward output layer. The network outputs the predicted target of the wrist $\tve{p}_f(t)$ at time $t$.}
    \label{fig:lstm}
\end{figure*}


\subsection{Target prediction Model}
\label{sec:target_position}

Prediction based on acceleration data requires the integration of the signals along with position information. However, the motion is non-linear and an analytical representation is not available. Consequently and in addition to the sequential nature of the data, we require learning the motion pattern over some period. As noted above, LSTM has the ability to model sequences by selectively remembering certain patterns over some period and learn long-term dependencies. Therefore, we utilize LSTM to explore the various feature and data size requirements in order to predict targets.

Let $S_H(t)\in\mathbb{R}^n\times\ldots\times\mathbb{R}^n$ be the sequence of $H$ past states up to time $t$ given by 
\begin{equation*}
    S_H(t)=\{\ve{x}(t-H),\ldots,\ve{x}(t-1),\ve{x}(t)\}.
\end{equation*}
From each recorded episode $\{\mathcal{X}_j,\mathcal{P}_j\}$, we extract sequences $S_H(H),S_H(H+1)\ldots S_H(t_f)$ and their respected target label $\ve{p}(t_f)$. Label $\ve{p}(t_f)$ is the last component of $\mathcal{P}_j$. Consequently, we acquire a training dataset from all episodes $\mathcal{L}=\{(S_{H,i},\ve{p}_{f,i})\}_{i=1}^M$ where $S_{H,i}$ and $\ve{p}_{f,i}\in\mathcal{W}$ are the $i^{th}$ sequence and corresponding target label, respectively, in the dataset.

Using dataset $\mathcal{L}$, we can directly train an LSTM model for map $\Phi$ to predict wrist target. In such case, the input to the LSTM would be a sequence of states where each state has a maximum dimension of $n=18$. In the experimental section, we will further investigate the importance of IMU features to the accuracy of the prediction. Nevertheless, we hypothesize that the corresponding estimation (using trained model $\Gamma_\theta$) of wrist positions included in the state sequence would provide viable information for better accuracy. This is analogous to including initial conditions when integrating acceleration. Hence, we consider two alternatives for a new state input $\bve{x}(t-k)$ to the LSTM. In the first, we would feed only the approximated wrist positions to the LSTM
\begin{equation}
    \label{eq:bvex}
    \bve{x}(t-k)=\Gamma_\theta(\ve{x}(t-k))^T.
\end{equation}
Alternatively, we observe the concatenation of the wrist position and the original raw data. Hence, 
for each state in input sequence $S_H(t)$, we concatenate the approximated wrist position to generate a new state $\bve{x}(t-j)$ given by
\begin{equation}
    \label{eq:concat}
    \bve{x}(t-k)=\left(\ve{x}^T(t-k),\Gamma_\theta(\ve{x}(t-k))^T\right)^T.
\end{equation}
Consequently, the input to the LSTM will now be the sequence along with the corresponding approximated wrist position. We would analyze these two alternatives in the experimental section.

We denote LSTM with wrist position information as \textit{LSTM-Pos}. The architecture of an LSTM-Pos layer is seen in Figure \ref{fig:lstm}a. The original state sequence $S_H(t)$ is the input to the layer followed by concatenation \eqref{eq:concat} using trained model $\Gamma_\theta$. When considering state representation \eqref{eq:bvex}, the concatenation does not occur and approximated wrist positions are fed directly into the cells. A cell in the layer receives vector $\bve{x}(t-k)$ and outputs hidden state vector $\ve{h}(t-k)\in\mathbb{R}^m$ where $m$ is an hyper-parameter. Hidden state vectors are passed between the cells and additionally collected to hidden state sequence $D(t)=\{\ve{h}(t-H),\ldots,\ve{h}(t)\}$. Sequence $D(t)$ is the output of the LSTM-Pos layer.

Figure \ref{fig:lstm}b illustrates the entire LSTM-Pos network. LSTM-Pos has $b$ rows while each row has $a$ layers. $a$ and $b$ are hyper-parameters of the network. Only the layers of the first row are LSTM-Pos layers where each outputs an hidden state sequence $D_j(t)$ for $j=1,\ldots,a$. The other layers are regular LSTM layers that receive hidden state sequences and output updated ones. While all layers in a row have similar architecture, they include a random dropout such that $D_i(t)\neq D_j(t)$ for any $i,j\in\{1,\ldots,a\}$ and $i\neq j$. Furthermore, only hidden states at time $t$ are outputted from the last row. Therefore, $a$ hidden states $\ve{h}_1(t),\ldots,\ve{h}_a(t)$ are used and are the input to a standard Convolutional Neural-network (CNN) and further to a fully-connected NN (FC-NN). The FC-NN outputs the approximated target wrist position $\tve{p}_f(t)$ predicted at time $t$. In preliminary testings, training the model without a CNN failed to converge and is, therefore, essential. In the experimental section, we compare the LSTM-Pos to the standard LSTM which has the same architecture while not including wrist position information.


\subsection{Curriculum learning}

Preliminary testing has shown that training the models directly with all data may lead to predictions converging to the geometric center of the workspace. Hence, we use \textit{Curriculum Learning } (CL) \cite{Wang2021}. The common approach is to introduce the model with data in an organized order from easy samples to hard ones. Such approach commonly provides better performance than random data shuffling. 

We use preliminary insights on model performance to tackle the problem of convergence to a local minima. We propose to take a reverse CL approach and train with harder samples first. In early studies, we have noticed that it is harder to predict the wrist pose and target at early stages of the reaching motion. This is due to low data variance when initiating motion. Hence, we train a model with $N_{cl}$ batches along the motion starting from time $t=0$. Once the loss value of the model successfully reaches below some user-defined value $\gamma_{cl}$, the next batch of further time frame is added to the training. When optimizing the hyper-parameters of some model, if value $\gamma_{cl}$ is not reached for some batch, the model is disqualified.


%% file: experiments.tex
\begin{figure*}
    \centering
    \includegraphics[width=\linewidth]{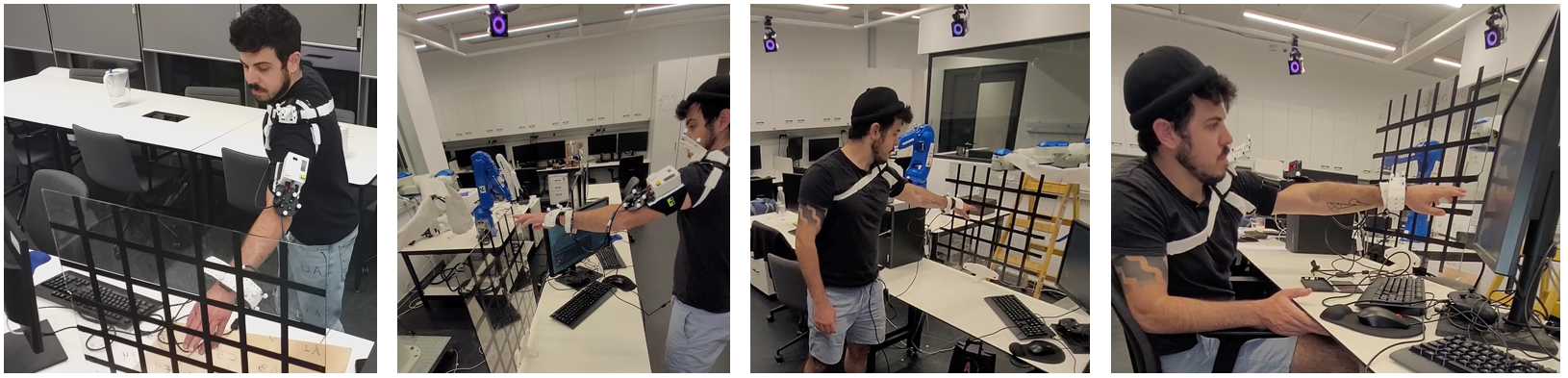}
    \caption{Participant 1 collecting data of arm reaching in various torso orientations including facing the collection board, body perpendicular to the board and while sitting down.} 
    \label{fig:collection}
\end{figure*}
We use the experimental system described in Section \ref{sec:system} to collect data and analyze the proposed approach. Collection was performed by one human participant (participant 1) with the system on his left arm. Arm lengths can be seen in Table \ref{tb:arm_lengths}. To achieve uniform and guided data collection, a rectangular collection board was mounted in front of the participant as seen in Figure \ref{fig:collection}. The board was equally divided into 42 squares. The participant was asked to start from arbitrary arm locations near the body and reach each of the squares several times while touching random locations within each square. To acquire a robust model, the participant also removed the system and re-positioned it on the arm several times during recordings. It is important to note that the proposed method processes raw data and uses relative IMU signals. Therefore, calibration of the IMU sensors is not required and the device can be used immediately upon positioning.

\begin{figure}
    \centering
    \includegraphics[width=\linewidth]{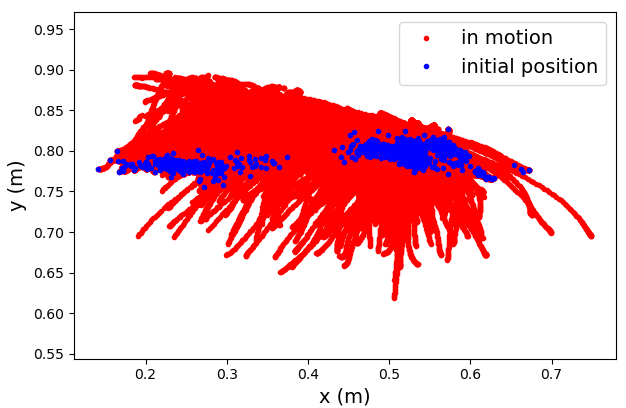}
    \caption{Shoulder positions relative to the board (at $y=0$) before initiating motion (blue) and during motions (red).} 
    \label{fig:shoulder_perturbations}
\end{figure}

A total of $K=840$ reaching episodes were collected with 20 episodes for each square. Preliminary analysis has shown that natural reaching motions in the described setup are no longer than 2 seconds. Hence, each episode was recorded in 60~Hz for $T=2~sec$ leading to 120 samples for an episode. Therefore, a set of $N=100,800$ samples are available for training a wrist position model. The participant recorded reaching tasks with various torso orientations including facing the collection board, body perpendicular to the board and while sitting down. Hence, a large distribution of shoulder motions during arm reaching can be seen in Figure \ref{fig:shoulder_perturbations} showing high variance perturbations. Note however that explicit information about the pose of the human during reaching is not included in the training data.

A large testing dataset was also collected independent of the training set and includes 336 episodes when reaching eight episodes for each square in various torso poses. The test set also included removing and re-positioning the device between episodes in order to test robustness. We next evaluate the training wrist position and target prediction models with the data. The datasets generated during and analysed during the current study are available in a designated repository\footnote{Dataset available at: \texttt{https://bit.ly/3COveC8}}

\begin{table}[]
\centering
\caption{Approximation accuracy of wrist position during reaching motion}
\label{tb:wrist_position}
\begin{tabular}{|l||c|c|}\hline
& w/o CL & w/ CL \\
&  (mm) &  (mm)\\\hline
All sensors    & \cellcolor[HTML]{C0C0C0}58.24 $\pm$ 4.79    & \cellcolor[HTML]{C0C0C0}48.62 $\pm$ 5.07 \\
Accelerometers & 99.71 $\pm$ 18.43   &  63.78 $\pm$ 14.73\\
Accel. \& Magn. & 76.1 $\pm$ 5.63     & 58.81 $\pm$ 5.83 \\
Gyroscopes     & 144.08 $\pm$ 17.24  & 124.79 $\pm$ 20.32 \\
Magnetometers  & 93.79 $\pm$ 11.87   &  65.55 $\pm$ 8.43\\\hline
Wrist only     & 64.51 $\pm$ 5.19    & 55.76 $\pm$  5.86\\
Upper-arm only & 149.1 $\pm$ 14.39   & 59.78 $\pm$ 10.8\\\hline
\end{tabular}
\end{table}

\subsection{Wrist position prediction evaluation}

We now analyse the wrist position model $\Gamma$ as discussed in Section \ref{sec:wrist_position}. We analyze the accuracy of various ANN models trained with different features in the data. We test the ability of a model to predict the wrist position if only some of the features are available including: only accelerometers, only magnetometers, only gyroscopes, only accelerometers and magnetometers, only wrist IMU band, only upper-arm IMU band or all available sensors. For each variation, we optimized the hyper-parameters of the ANN to minimize the RMSE loss function along with the Adam optimizer. For example, the optimal architecture for the model with all sensing features included three hidden layers with 512 neurons each and a rectified linear unit (ReLU) activation function. We also compare between direct and CL training of the ANN. For the CL, we manually optimized the batch size and loss threshold yielding $N_{cl}=10$ and $\gamma_{cl}=58~mm$, respectively.
\begin{figure}
    \centering
    \includegraphics[width=\linewidth]{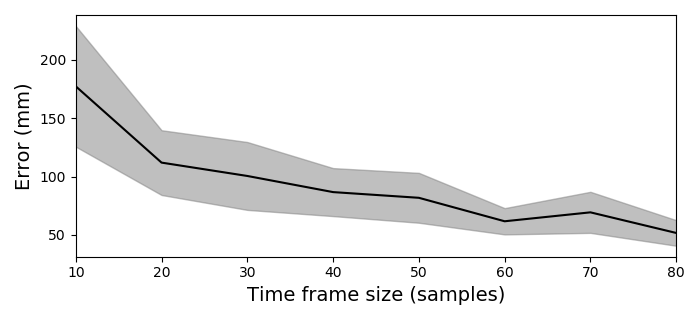}
    \caption{Target prediction error (mean and standard deviation) with respect to the number of the past states $H$.} 
    \label{fig:Error_window}
\end{figure}
\begin{table*}[]
\centering
\caption{Target prediction accuracy for LSTM and LSTM-Pos}
\label{tb:Target_pred}
\begin{tabular}{|l|cc|cc|}\hline
Features       & \multicolumn{2}{c|}{LSTM} & \multicolumn{2}{c|}{LSTM-Pos} \\
               & Mean (mm) & Std. (mm)     & Mean (mm) & Std. (mm) \\\hline
All sensors            & 194.63 & 73.07 & 61.46 & 11.26 \\
Accelerometers         & 195.42 & 70.03 & 66.27 & 12.42 \\
Accel. and Magn.       & 195.37 & 69.96 & 65.53 & 12.56 \\
Gyroscopes             & 195.41 & 70.02 & 71.58 & 14.04 \\
Magnetometers          & 195.43 & 69.99 & 72.71 & 19.12 \\
Approx. wrist position & -      & -     & \cellcolor[HTML]{C0C0C0}59.99 & \cellcolor[HTML]{C0C0C0}12.01 \\\hline
Wrist only             & 195.39 & 69.99 & 67.56 & 13.9  \\
Upper-arm only         & 195.42 & 70.03 & 69.04 & 17.26 \\\hline
\end{tabular}
\end{table*}

Table \ref{tb:wrist_position} summarizes the accuracy results for all variations. It is clear that CL provides accuracy improvement in all feature representations. Magnetometers (denoted \textit{Magn.} in the table) by themselves supposedly should provide enough information to localize the arm. However and due to variations in torso orientations, moderate accuracy is provided with only magnetometers. Only using accelerometers (denoted \textit{Accel.} in the table) reaches similar accuracy as they are able to encapsulate motion flow patterns but lack orientation information. Evidently, combining accelerometers and magnetometers provides accuracy improvement. Furthermore, using all available sensors of both IMU's (including the gyroscopes) provides the lowest error. The results also show that having only one IMU either on the upper-arm or the wrist could provide fairly accurate wrist position. We note that these results are similar to results reported using the Unscented Kalman filter \cite{Atrsaei2018}.   

\subsection{Target prediction evaluation}

With the above NN models for approximating wrist position, we analyze the prediction of the target in reaching motions with the LSTM-Pos. We compare LSTM-Pos to an LSTM without including explicit approximation of wrist positions. Here also, we analyze the required features necessary to achieve accurate prediction. The hyper-parameters of models with different feature variations were optimized to yield the lowest RMSE loss value. For example, the optimal hyper-parameters for the model with all sensing features include LSTM with $a=256$ and $b=2$, $m=64$, CNN with three convolutional layers and FC-NN with one hidden layer of 14 nodes, yielding 42,291 trainable parameters. In this part, all training is performed with CL since it provides better learning as demonstrated previously.

We initially observed the required number of past states $H$ to acquire an accurate model. Figure \ref{fig:Error_window} shows the mean prediction error over the test data when including all sensors and with regards to $H$. Including more information provides, as one would expect, better accuracy. Other feature variations exhibit similar behaviour. Having a large $H$ means that the first prediction would arrive later in the reaching motion. Hence, we face a trade-off between earlier prediction and accuracy. In further analysis we choose to use $H=60$ as it provides low error along with first prediction at half-way through-out the reaching motion. The chosen prediction horizon provides at least one second before the hand reaches the target which is valuable for planning and initiating robot motion. While not implemented in this work, these results show that one could train several models with ascending $H$. Hence, coarse accuracy would be provided at the beginning of the episode (e.g., $H=10$) for initial motion of the robot. Then, accuracy would be improved with better models as more information arrive.

\begin{figure}
    \centering
    \includegraphics[width=\linewidth]{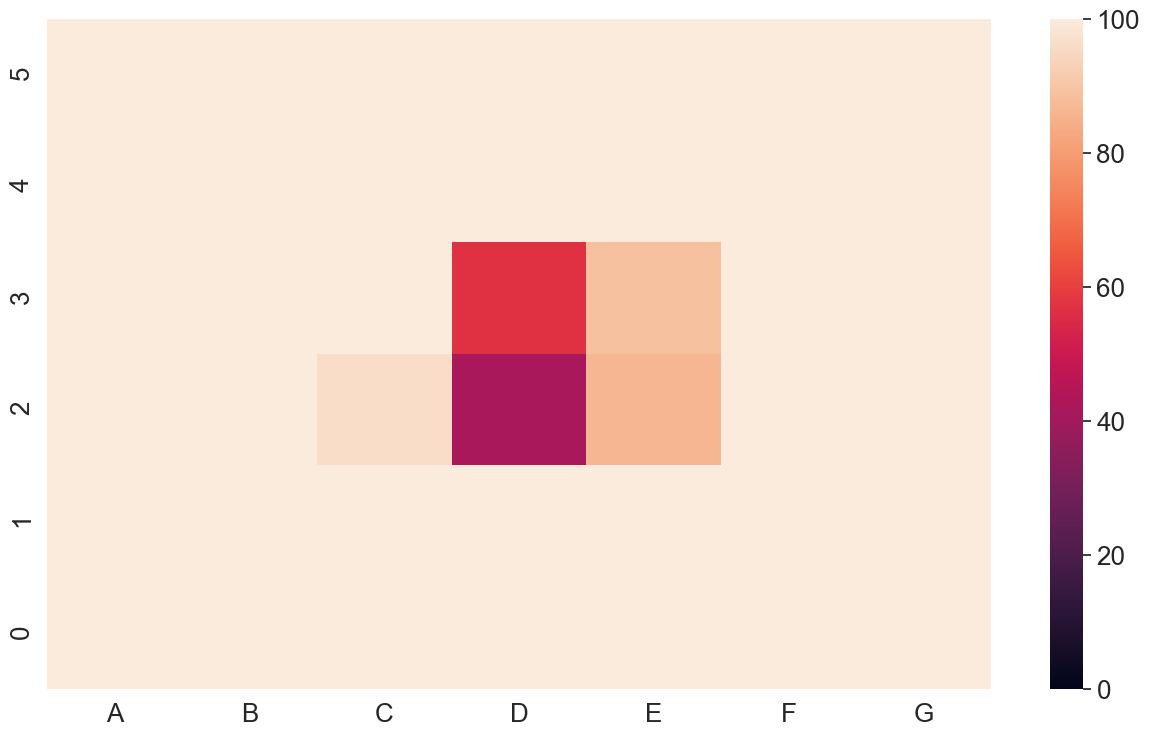}
    \caption{Typical Heat-map illustrating the mean error (in \textit{mm}) across the collection board when not including wrist position approximation of LSTM-Pos. All models converged their prediction to the center of the board. } 
    \label{fig:hm_no_mlp}
\end{figure}

Table \ref{tb:Target_pred} summarizes the target prediction errors (mean and standard deviation) along the reaching motion of all test data and for different feature variations. First, directly using the data to train an LSTM (without including wrist position approximation) fails to produce feasible predictions. All LSTM models converged to the mean of the target labels. Figure \ref{fig:hm_no_mlp} shows an heatmap illustrating the error distribution across the collection board. Predictions for all test episodes output the center position of the collection board leading to low accuracy performance. 

\begin{figure}
    \centering
    \includegraphics[width=\linewidth]{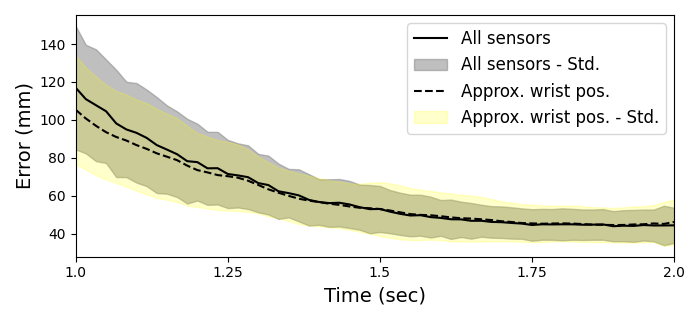}
    \caption{Target prediction error (mean and standard deviation) of LSTM-Pos with respect to reaching motion time evaluated on all test episodes.} 
    \label{fig:Error_time}
\end{figure}

\begin{figure*}
    \centering
    \includegraphics[width=\linewidth]{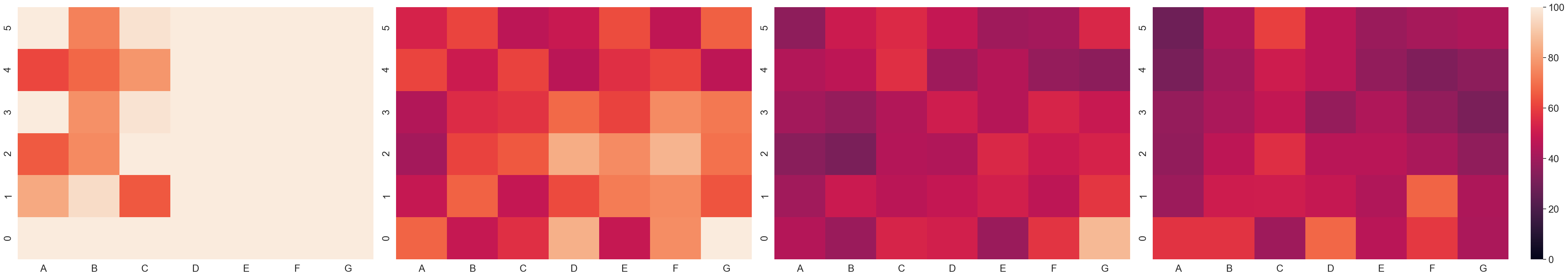}
    \caption{Heat-maps illustrating the mean error (in \textit{mm}) across the collection board over time and when considering all sensors. Mean error from left to right: $Error(t=1~sec)=116.98~mm$, $Error(t=1.33~sec)=62.41~mm$, $Error(t=1.65~sec)=46.86~mm$ and $Error(t=2~sec)=44.47~mm$. } 
    \label{fig:error_heatmap}
\end{figure*}

LSTM-Pos, on the other hand, provides much lower prediction errors for all feature variations. In particular, having only the sequence of approximated wrist positions from the NN fed into the LSTM is sufficient to acquire the best model. Including also the raw measurements from all sensors fairly provides the same accuracy. The results also indicate that having only one band can be sufficient for moderate accuracy. Figure \ref{fig:Error_time} shows the mean error along the test episodes for LSTM-Pos with only approximated wrist positions and when including also raw sensor measurements (all sensors). Similarly, Figure \ref{fig:error_heatmap} illustrates the mean errors across the collection board through the motion time. The above results achieve accuracy of approximately less than 70\% of an adult human palm breadth (87.5~mm) \cite{Bayraktar2018} and is sufficient for feasible HRC tasks as would be demonstrated in the next section. 

\subsection{HRC Demonstration}

We have conducted an HRC demonstration where a participant reaches his arm towards a robotic arm. During the reaching motion, the robotic arm receives continuous stream of target predictions acquired from the trained model. When acquiring the first prediction, the robot plans motion for rendezvous and initiates motion. As a new prediction arrives, the motion plan of the robot is updated accordingly. Participant 1 along with two additional ones not included in training participated in the demonstration. The participants have different arm lengths as seen in Table \ref{tb:arm_lengths}. They were instructed to stand in front of the robot at desired body poses such that their wrist can reach the shared workspace. These poses are roughly similar to the ones used for collecting training data and within the same distribution. Furthermore, the robot was not given any information regarding the pose of the human participant. The participants were asked to perform 15 reaching motions to arbitrary locations in front of them. They were also instructed to initiate motion from any desired arm pose. We define a successful trial as a one where the robot reached within the vicinity of the human hand with minimal distance of 60~mm. The distance was measured using the motion capture system. 

\begin{table}[]
\centering
\caption{Arm lengths of all participants}
\label{tb:arm_lengths}
\begin{tabular}{|c|ccc|}\hline
Participant   & 1 & 2 & 3           \\\hline
$l_u$ (cm)    & 29   & 33.5 & 25    \\
$l_f$ (cm)    & 28.5 & 30.3 & 24.7  \\\hline
\end{tabular}
\end{table}

Table \ref{tb:reaching} provides the rendezvous success rate for the three participants. The success for Participant 1 is the highest since the model is trained on data collected from him. Nevertheless, the two other participants achieved rather high success rates considering that they have different arm lengths and did not contribute data to model training. Hence, the model can be transferred to a new user with relatively good accuracy. While the participants were not instructed to stand in a specific pose, yet the model was able to provide accurate predictions. This can be explained by the ability of the model to infer from motion patterns, not only regarding the trajectory of the arm, but also about the human pose. In other words, information regarding human pose is embedded within state trajectories and the model can cope with pose variance for successful HRC. Hence, the model is shown to provide good accuracy within some implicit pose distributions of the human participant.

\begin{figure*}
    \centering
    \includegraphics[width=\linewidth]{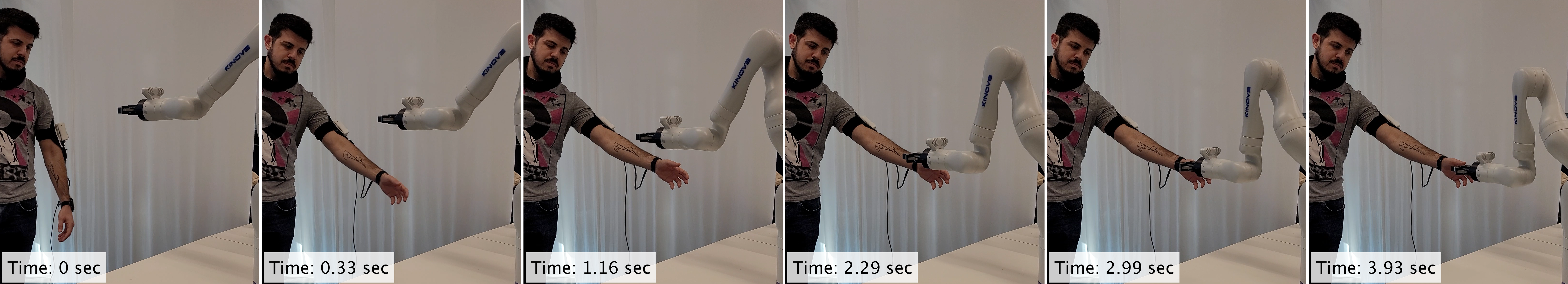}
    \caption{Reaching test of Participant 1.} 
    \label{fig:N_sequence1}
\end{figure*}

\begin{figure*}
    \centering
    \includegraphics[width=\linewidth]{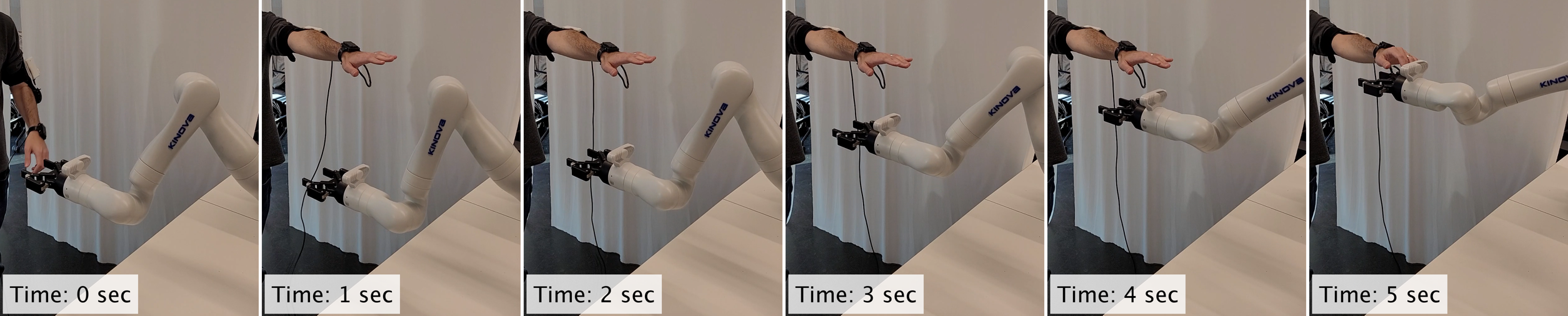}
    \caption{Reaching test of Participant 3.} 
    \label{fig:A_sequence1}
\end{figure*}

\begin{figure}
    \centering
    \includegraphics[width=\linewidth]{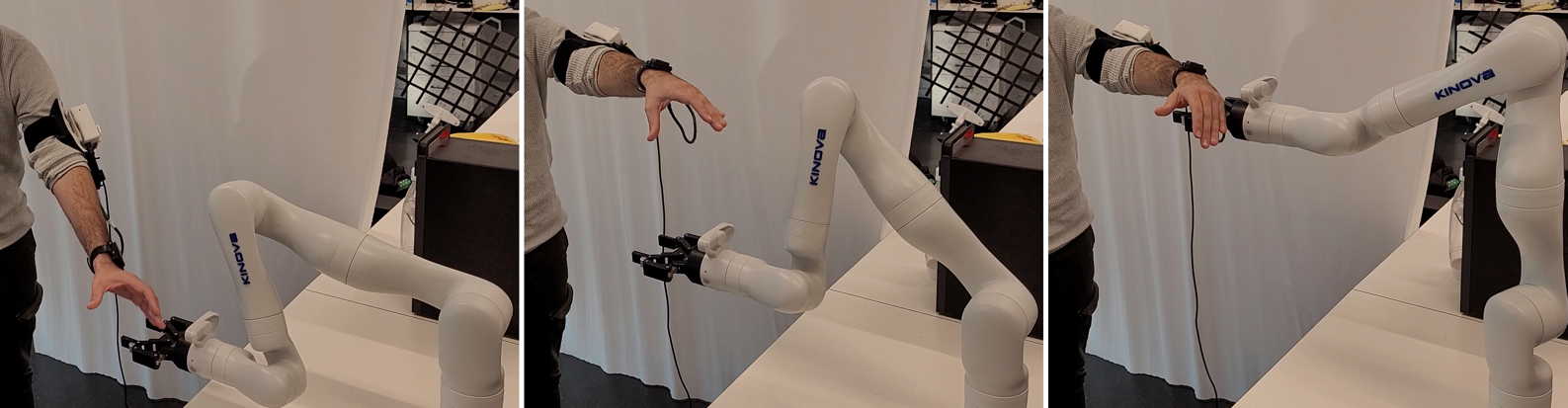}
    \caption{Motion of participant 2 between two arm poses within the interaction space while the robotic arm follows.} 
    \label{fig:Noa_sequence1}
\end{figure}

\begin{figure}
    \centering
    \includegraphics[width=\linewidth]{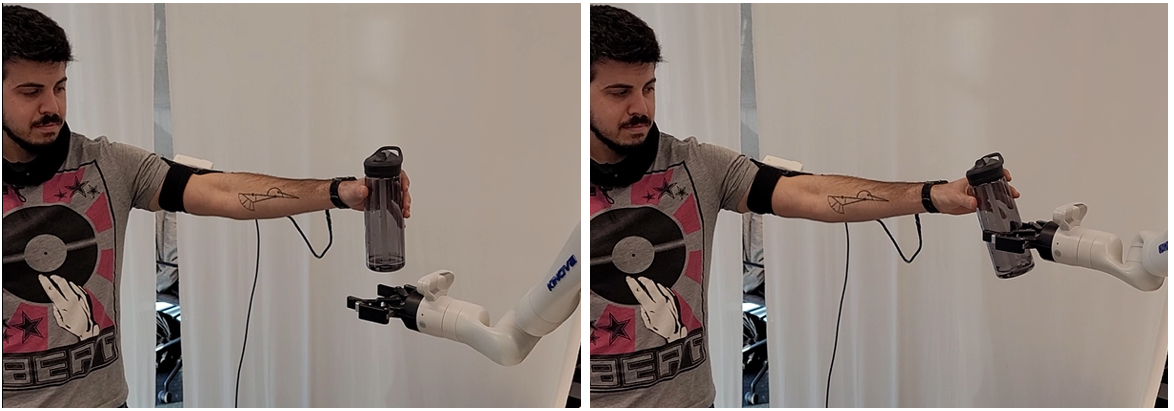}
    \caption{An HRC demonstration in which participant 1 hands-over a bottle to the robot.} 
    \label{fig:bottle}
\end{figure}

We note that the robot arm reached the vicinity of the human hand in all trials but with a distance larger than the defined bound. Target wrist prediction was achieved in mid-motion and the robot initiated motion before the participant reached the end of the episode. For safety reasons, the velocity of the robot was maintained low. Hence, the participant was required to shortly wait for the arm. Nevertheless, having the robot move faster would enable faster HRC tasks. Figures \ref{fig:N_sequence1}-\ref{fig:Noa_sequence1} show demonstration snapshots of some reaching trials for the three participants towards the robotic arm. Participant 3 in Figure \ref{fig:Noa_sequence1} demonstrated the ability of the model to predict motion from various initial poses of the arm. Figure \ref{fig:bottle} shows a demonstration of handing over a bottle to the robot using the system. These results show that a relatively small dataset of IMU data can be used to train a sufficiently accurate model for feasible predictions of reaching targets in HRC scenarios. All experiments and demonstration videos can be seen in the supplementary video.

\begin{table}[]
\centering
\caption{Success rate for reaching experiment}
\label{tb:reaching}
\begin{tabular}{|c|c|}\hline
Participant             & Success rate \\\hline
1    & 93.3\%      \\
2    & 80.0\%      \\
3    & 86.6\%      \\\hline
\end{tabular}
\end{table}


%% file: conclusions.tex
We have presented the problem of learning human reaching motions and predicting future target locations. We investigated the use of a wearable device composed of two IMU senors located on the upper-arm and wrist. First, a simple ANN was trained to predict the current position of the wrist. Then, an LSTM-based architecture was proposed where approximation of the wrist position along some time frame was included. Using data collected from a single user, we have tested various features to be included in the model. Results show that having all features from the sensors along with curriculum learning achieve good accuracy for approximating current wrist position. With such model and without additional raw data, we reach sufficient accuracy for early prediction of the target wrist position. Furthermore, we have shown several reaching demonstrations where a robot planned and moved towards the human arm in real-time based solely on information from the wearable device. High success rates were demonstrated from the participant whom collected data and two other ones not included in the training. Hence, a relatively small amount of collected data from one user achieved sufficient accuracy for feasible predictions of reaching targets in HRC scenarios. Once trained, the model and system can be deployed in various spaces with no further effort.

Future work could focus on calibrating the existing model to a new user based on a limited amount of collected data. The system could also be fused with additional sensors. For instance, an additional IMU can be added on the human torso to provide information regarding body orientation and upper body shift. Vision can also be integrated in order to provide a complete solutions when a line-of-sight is not continuous. Similarly, one could integrate Force-Myography \cite{kahanowich2021,Bamani2022} or EMG \cite{Bi2019} to have additional information about the pose of the palm and fingers. An additional system on the other arm could enable the classification of tasks performed by the human towards assistance by a robot.